\documentclass{article}
\usepackage{spconf,amsmath,amssymb,graphicx}
\usepackage[table]{xcolor}
\usepackage{booktabs}
\usepackage{float}
\usepackage[hidelinks]{hyperref}
\usepackage{bbm}
\usepackage{cite} 
\usepackage{tcolorbox}
\tcbuselibrary{skins}
\definecolor{rqframe}{HTML}{0000FF}
\definecolor{rqfill}{HTML}{C6F1FF}
\newtcolorbox{rqbox}{enhanced, colframe=rqframe, colback=white, boxrule=0.8pt,
  arc=3mm, left=2mm, right=2mm, top=0.5mm, bottom=0.5mm, before skip=2pt, after skip=3pt}
\definecolor{hallucT}{HTML}{B35A00}
\definecolor{groundedT}{HTML}{4C8A00}
\definecolor{eostokT}{HTML}{6B21A8}
\definecolor{grouprow}{gray}{0.91}
\definecolor{ctxgray}{gray}{0.50}

\newsavebox{\slamfig}

\title{Induction Head Origins of SLAM-ASR Speech Recognition Models}
\name{Artem Fedorchenko\sthanks{Corresponding author: artem.fedorchenko@taltech.ee}, Tanel Alum\"{a}e}
\address{Tallinn University of Technology, Estonia}
\begin{document}
\ninept
\maketitle
\begin{abstract}
What makes a small trained map from encoder states to the embedding space of a frozen language model sufficient for competitive speech recognition? The premise of the SLAM-ASR architecture is that a frozen language model possesses linguistic knowledge rich enough to perform the recognition task. This assumption, while reasonable, provides little insight into the mechanisms underlying these models. We therefore seek the concrete internal structures on which they rely. This work shows that SLAM-ASR models rely on induction heads of the frozen language model: the same heads that copy repeated token sequences in text are also important for mapping speech embeddings to transcript tokens. Moreover, we find pairs of heads that behave as a constrained form of differential attention, cancelling attention noise and sharpening attention onto the previously generated word. Finally, we show how induction heads could be used to improve transcription quality under noisy audio. 
\end{abstract}

\begin{keywords}
speech recognition, SLAM-ASR, hallucination
\end{keywords}
\section{Introduction}
\label{sec:intro}
\suppressfloats[t]
\begin{figure}[t]
  \centering
  \includegraphics[width=\columnwidth]{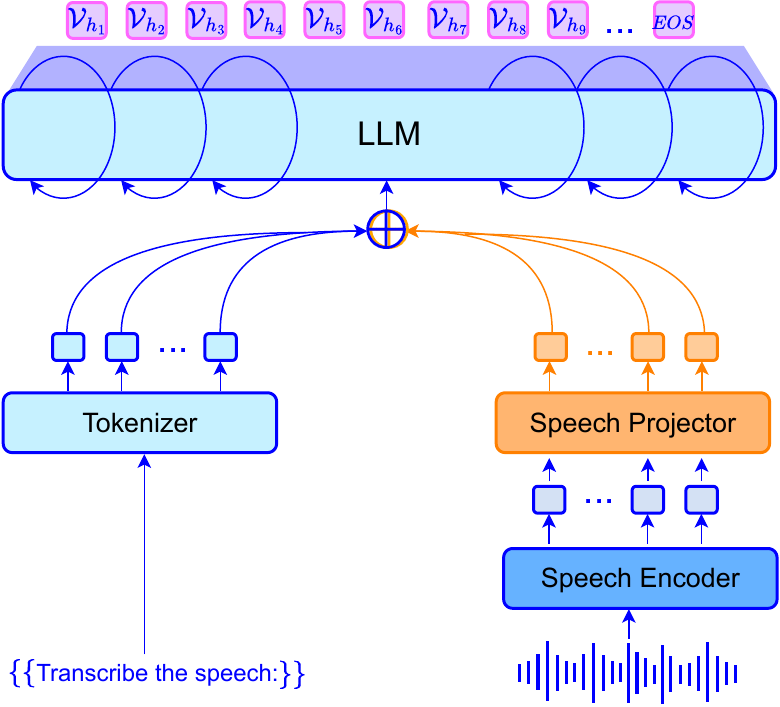}
  \caption{SLAM-ASR architecture. \textcolor[HTML]{FF8000}{Speech projector} is the only component being trained, while \textcolor[HTML]{0000FF}{LLM} and \textcolor[HTML]{0000FF}{speech encoder} remain frozen.}
  \label{fig:slam-asr}
\end{figure}
\begin{figure*}[!t]
  \centering
  \includegraphics[width=\textwidth]{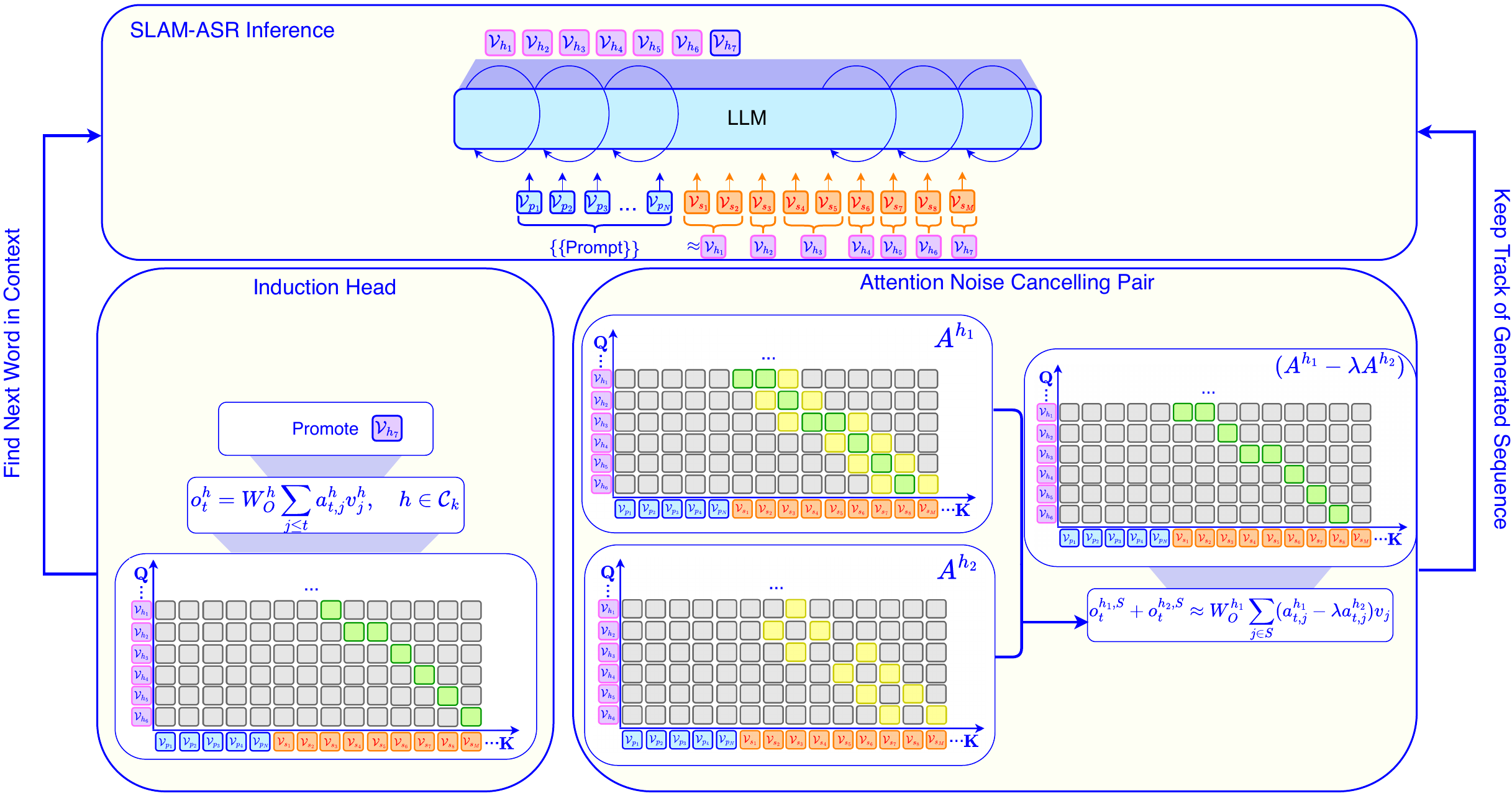}
  \caption{Per-head contribution to generated sequence at token $\textcolor{blue}{\mathcal{V}_{h_i}}$, which predicts $\textcolor{blue}{\mathcal{V}_{h_{i+1}}}$. \textbf{Bottom left:} induction heads attend to the speech embeddings of $\textcolor{blue}{\mathcal{V}_{h_{i+1}}}$. \textbf{Bottom right:} an attention noise cancelling pair jointly sharpens attention on the speech embeddings of $\textcolor{blue}{\mathcal{V}_{h_i}}$.}
  \label{fig:heads}
\end{figure*}

Integrating features representing the input audio with language models is at the core of automatic speech recognition (ASR)~\cite{jelinek1976}. Recently, both of these constituents improved vastly, but separately, largely due to the widespread use of the Transformer architecture~\cite{vaswani2017}. Speech encoder and language model fusion approaches were suggested in~\cite{karita19_interspeech,gulcehre2015using,sriram18_interspeech}; see~\cite{schmitt2026llms} for details. 
An alternative method is to use a general-purpose large language model (LLM) to generate a corrected transcript for an ASR hypothesis \cite{yang2023generative, fedorchenko2025}. However, the most popular method in recent years, commonly referred to as speech LLM (SLLM), is to train an LLM to also accept acoustic information from an audio encoder as input, through a learnable speech projector. 

Early works of this kind~\cite{wu2023,fathullah2023,chen2023} trained the encoder and adapted the LLM through low-rank updates, together with the speech projector. SLAM-ASR~\cite{slamasr} (Fig.~\ref{fig:slam-asr}), however, trains only the projector, with both encoder and LLM being frozen, yet still achieves competitive recognition. The outputs of such a projector need not coincide with token embeddings, and its sequence length differs from the text counterpart, so linguistic modelling and alignment between two sequences are to be solved by the LLM. The premise of SLAM-ASR, that the frozen LLM suffices for the linguistic task, does not name the mechanisms on which it relies, in particular, what makes the frozen decoder capable of reading the projected speech embeddings.

Previous studies in internal analysis of SLLMs~\cite{hsu2026} found that in a spoken question answering context, speech and text states are similar in a broad band of middle layers. In~\cite{glazer_beyond}, the authors applied mechanistic interpretability tools to ASR models to analyze encoder representations and error modes such as repetition hallucination, which they linked to a single attention head in Whisper~\cite{whisper}. In a frozen-decoder setting like ours, many decoder layers were found redundant for ASR~\cite{moumen_redundancy}, which suggests that recognition capability is concentrated, possibly at the head level. In this vein, text-to-speech in-context learning (ICL) work~\cite{pouw_icl} considered the role of induction heads~\cite{elhage2021,olsson2022}, showing that, as in text, ablating them hurts ICL more than ablating random heads. In the audio LLM domain, \cite{das_inherited} considered using text-backbone heads to steer speaker attribution. Another work~\cite{guo_ahamask} learned head masks to specify tasks, including ASR. For the frozen LLM decoder scenario as in SLAM-ASR, to the best of our knowledge, per-head impact on ASR has not been studied. Building upon the soft-prompting view~\cite{lester_prompt,li_prefix}, in which the decoder is presented with a noisy and redundant version of the embedded transcript, we consider the following research questions (RQs):

\par\vspace{5.5pt}\noindent\textbf{(RQ1)} \emph{Does ASR quality depend on induction heads present in the frozen LLM?}\par\noindent\textbf{(RQ2)} \emph{Are there heads responsible for mitigating redundancy of speech embeddings?}\par\noindent\textbf{(RQ3)} \emph{Does scaling induction heads at inference help recover recognition quality under noisy audio?}\par\vspace{5.5pt}
We show that: (1) ASR in SLAM-ASR models depends on \textbf{induction heads} of the frozen LLM, (2) the decoder contains pairs of heads that act together as a constrained form of \textbf{differential attention}~\cite{ye2025}, sharpening attention on the speech embeddings of the previously generated word; and (3) scaling heads of (1) at inference reduces WER and insertions under noisy audio in 2 of 3 LLMs.

\section{Background}
\label{sec:background}
This work connects SLAM-ASR speech recognition models with induction heads and differential attention. We therefore give a brief intuition behind these two notions; for a more complete picture, see~\cite{elhage2021} and~\cite{ye2025}, respectively.

\textbf{Induction heads.} On a sequence of repeated tokens $\dots A\,B \dots\allowbreak A$, an induction head is an attention head that, at the second occurrence of $A$, attends to the token $B$ that followed its first occurrence (\emph{prefix matching}) and raises the logit of $B$ (\emph{copying}), so that the model continues the sequence with $B$~\cite{elhage2021,olsson2022}. Mechanistically, a previous-token head in an earlier layer writes the token that precedes each position into it, so the key $\textcolor{blue}{\mathbf{K}}$ at $B$ carries ``preceded by $A$'' information, which the query $\textcolor{blue}{\mathbf{Q}}$ at the second occurrence of $A$ matches. Such heads emerge during text pretraining of an LLM, together with ICL~\cite{olsson2022}, and can be found in a trained model via the prefix-matching score (Sec.~\ref{sec:methodology}). In our context, the speech embeddings play the role of the first occurrence (Fig.~\ref{fig:heads}): the speech embeddings $\textcolor{red}{\mathcal{V}_{s_j}}$ of each word approximate its generated token $\textcolor{blue}{\mathcal{V}_{h_i}}$, e.g., $\textcolor{red}{\mathcal{V}_{s_1}}, \textcolor{red}{\mathcal{V}_{s_2}} \approx \textcolor{blue}{\mathcal{V}_{h_1}}$, and are followed by those of the next word. At a generated token $\textcolor{blue}{\mathcal{V}_{h_i}}$, the second occurrence, an induction head therefore attends to the speech embeddings approximating $\textcolor{blue}{\mathcal{V}_{h_{i+1}}}$, which is the pattern shown in the bottom left of Fig.~\ref{fig:heads}.

\textbf{Differential attention.} Differential attention, proposed in~\cite{ye2025}, is an enhancement of the canonical Transformer, which mitigates attention noise, i.e., the non-negligible attention that softmax assigns to irrelevant context. Differential attention computes two softmax attention maps and subtracts the second, scaled by $\lambda$, from the first:
\begin{equation}
  \big(\mathrm{softmax}(\textcolor{blue}{\mathbf{Q}}_1\textcolor{blue}{\mathbf{K}}_1^{\top}/\sqrt{d}) - \lambda\, \mathrm{softmax}(\textcolor{blue}{\mathbf{Q}}_2\textcolor{blue}{\mathbf{K}}_2^{\top}/\sqrt{d})\big)\textcolor{blue}{\mathbf{V}}.
\end{equation} 
As in a differential amplifier, whose output is proportional to the difference of its two inputs, $G\,(v_{+} - v_{-})$, so that the signal common to both is rejected~\cite{ye2025}, attention that both maps assign to the same positions cancels, and the resulting attention is sharper on the relevant context. In~\cite{ye2025}, the two maps and $\lambda$ are built into the architecture and trained from scratch; a pretrained LLM has no such component. In our context, attention noise can be expected to grow with the redundancy of speech embeddings: each word spans several of them, so a head reading the speech sees more keys than for the same transcript as text, each receiving some attention. Two heads of the same layer can nevertheless act together as a constrained form of differential attention (Sec.~\ref{sec:methodology}); such a pair is shown in the bottom right of Fig.~\ref{fig:heads}, where its joint attention concentrates on the speech embeddings of the previously generated word, $\textcolor{blue}{\mathcal{V}_{h_i}}$.
\section{Methodology}
\label{sec:methodology}
\setlength{\abovedisplayskip}{2pt plus 2pt minus 1pt}\setlength{\belowdisplayskip}{2pt plus 2pt minus 1pt}
\subsection{Induction heads}  Denote by $(l,h)$ head $h$ of layer $l$, by $t$ and $j$ the query and key positions, by $q^{l,h}_t$, $k^{l,h}_j$, $v^{l,h}_j$ its query, key and value vectors, and by $S$ the set of positions of the speech embeddings, i.e., the outputs of the projector. The prefix-matching score of a head is computed on inputs $x = (\textsc{bos}, r_{1}, \dots, r_{N}, r_{1}, \dots, r_{N})$ with $N = 50$ and $r_i$ drawn uniformly from the vocabulary, so that the query at $t = N + i$ is the second occurrence of $r_i$:
\begin{equation}
  \mathrm{PM}(l,h) = \frac{1}{N-1}\sum_{t=N+1}^{2N-1} a^{l,h}_{t,\,t-N+1},
\end{equation}
where $a^{l,h}_{t,j} = \mathrm{softmax}_{j \le t}\big(q^{l,h\top}_{t} k^{l,h}_{j} / \sqrt{d}\big)$, averaged over 40 sequences. Let $\mathcal{C}_k$ be the set of the top-$k$ heads by $\mathrm{PM}$. As in~\cite{crosbie2025,pouw_icl}, we identify these as induction heads. The contribution of a set of heads to ASR is measured by setting their attention logits to $-\infty$ for $j \in S$, so that each head's attention is redistributed over the rest of the context, and by comparing the resulting increase in word error rate (WER) with that of random sets of $k$ heads drawn from the same layers, outside of $\mathcal{C}_k$.

\subsection{Attention noise cancelling} Two heads of the same layer that attend to the same speech embeddings but project what they read in opposite directions cancel part of each other's output. If the cancelled part is the attention that both heads assign to irrelevant positions, the pair as a whole attends more sharply to the relevant ones, as differential attention does (Sec.~\ref{sec:background}). Below, we state how such pairs are selected, the approximation under which a pair acts as one head whose attention is the difference of the two, and how we compare its sharpness with that of $h_1$ alone.

Without layer index, the output of head $h$ at position $t$ is $o^{h}_t = W_O^{h}\sum_{j \le t} a^{h}_{t,j} v^{h}_j$, where $W_O^{h}$ is the head's chunk of the output projection. Let $h_1$ be the head with the largest ASR importance, i.e., the largest increase in loss when only its logits for $j \in S$ are set to $-\infty$ (blocking), and $h_2$ the head of the same layer whose output is most anti-aligned with that of $h_1$. We consider such pairs (Fig.~\ref{fig:heads}) if both heads put on average at least half of their attention on $S$ and $h_2$ is far more anti-aligned with $h_1$ than any other head of the layer, measured by the mean of $\cos(o^{h_1}_t, o^{h_2}_t)$ over generated positions. In the pairs we find, the two heads share a key/value group of grouped-query attention (Sec.~\ref{sec:models}), $k^{h_1}_j = k^{h_2}_j$ and $v^{h_1}_j = v^{h_2}_j = v_j$, so they read the same values and differ only in where they attend, through their queries, and in how they project the result, through their output chunks. Denote by $o^{h,S}_t$ the output $o^{h}_t$ with the sum restricted to $j \in S$. The joint output of the pair on the speech embeddings is then, exactly,
\begin{equation}
  o^{h_1,S}_t + o^{h_2,S}_t = \sum_{j \in S,\, j \le t} \big(a^{h_1}_{t,j}\, W_O^{h_1} + a^{h_2}_{t,j}\, W_O^{h_2}\big) v_j .
\end{equation}
Anti-aligned outputs alone do not make this a subtraction of attention maps. Our reading rests on one approximation: $h_2$ projects each value in the direction opposite to $h_1$, $W_O^{h_2} v_j \approx -\lambda\, W_O^{h_1} v_j$, with
\begin{equation}
  \lambda = \operatorname*{arg\,min}_{\lambda'} \sum_{j} \big\lVert W_O^{h_2} v_j + \lambda' W_O^{h_1} v_j \big\rVert^2 ,
\end{equation}
where $j$ ranges over the positions in $S$ of all evaluation utterances; how well it holds is given by $\cos(W_O^{h_1}v_j, W_O^{h_2}v_j)$. Under it,
\begin{equation}
  o^{h_1,S}_t + o^{h_2,S}_t \approx W_O^{h_1} \sum_{j \in S,\, j \le t} \big(a^{h_1}_{t,j} - \lambda\, a^{h_2}_{t,j}\big) v_j ,
\end{equation}
i.e., the pair acts as one head with effective attention $a^{h_1}_{t,j} - \lambda\, a^{h_2}_{t,j}$: attention that $h_1$ and the $\lambda$-scaled $h_2$ assign alike to a position cancels. This is the form of differential attention~\cite{ye2025}, constrained in that the keys are shared, $\lambda$ is determined by the output chunks, and there is no per-head normalisation; in particular, the effective attention can be negative and need not sum to one. To measure what the subtraction does, let $T$ be the set of query positions at which a token of a word other than the first is generated, and $F_t \subseteq S$ the positions aligned to the previously generated word. We call the attention on $S \setminus F_t$ attention noise and measure the sharpness of an attention matrix $A$ by the share of its positive part that falls on $F_t$,
\begin{equation}
  \rho(A) = \frac{\sum_{t \in T}\sum_{j \in F_t}[A_{t,j}]_+}{\sum_{t \in T}\sum_{j \in S}[A_{t,j}]_+},
\end{equation}
where $[x]_+ = \max(x,0)$, comparing the pair's effective attention $A^{h_1} - \lambda A^{h_2}$ with that of $h_1$ alone, where $A^{h} = (a^{h}_{t,j})$.

\subsection{Inference-time head scaling} The performance of SLAM-ASR degrades significantly under additive noise~\cite{kumar2025}. To make use of $\mathcal{C}_k$, we consider a common ASR hallucination mode under such noise: an increased insertion rate $I$, the number of inserted words relative to the number of reference words. The decoder may then continue generating from its language prior; raising the output of heads that read the speech is expected to give the audio more weight against the prior. For each head $(l,h) \in \mathcal{C}_k$, we scale its output at every position, $o^{l,h}_t \mapsto \alpha\, o^{l,h}_t$, and compare the resulting WER and $I$ with those of the unscaled model, of equally many layer-matched random heads, and of all remaining heads of the same layers, each scaled by the same $\alpha$.

\section{Experiments}
\label{sec:experiments}
\subsection{Models}\label{sec:models}
We use the instruction-tuned Llama~3.2-3B~\cite{grattafiori2024}, Gemma~2-2B~\cite{gemmateam2024} and Ministral~3-3B~\cite{liu2026ministral} models as decoders; for Ministral, only the text decoder is used. All three decoders use grouped-query attention~\cite{ainslie2023}: the heads of a layer are split into groups, and the heads of a group share keys and values. The SLAM-ASR models use the Whisper turbo encoder~\cite{whisper}. The projector design follows~\cite{slamasr}: concatenate every $5$ consecutive encoder states and pass through a two-layer MLP with hidden size $2048$ and ReLU activation to the decoder's embedding space; we append a trainable end-of-speech embedding, initialised from the decoder's end-of-turn token. Encoder and decoder are frozen. The prompt $\textcolor{blue}{\mathcal{V}_{p_1}, \dots, \mathcal{V}_{p_N}}$ is ``Transcribe the speech:'' in each decoder's chat template.
\subsection{Datasets}
Each projector is trained on the \textit{train-clean-100} subset of LibriSpeech~\cite{librispeech} and validated on 500 utterances from \textit{dev-other}. Evaluation uses \textit{test-clean} and \textit{test-other}.
\subsection{Training Details}
Optimisation follows~\cite{slamasr}: AdamW~\cite{adamw} with learning rate $10^{-4}$ and no weight decay, $1{,}000$ warm-up steps followed by a constant learning rate, and early stopping with patience $3$, evaluated every $500$ steps. Unlike~\cite{slamasr}, we use batch size $32$ instead of $4$ and train for at most $5$ epochs instead of $100{,}000$ steps. Each decoder is trained with three seeds.
\subsection{Evaluation Protocol}
Decoding is done greedily. $h_1$ is selected by ASR importance on 200 \textit{dev-clean} utterances. Blocking and the head pairs are evaluated on 300 \textit{test-clean} utterances with Montreal Forced Aligner alignments~\cite{mcauliffe2017}: these give the word boundaries for $F_t$ and attention peaks. Each speech embedding covers 100\,ms and is assigned to the word whose boundaries contain its centre. ASR importance, head-pair quantities and attention peaks use teacher forcing. Blocking uses $k=10$ in all decoders. Scaling uses $\alpha=1.5$ and $k$ equal to 3\,\% of the decoder's heads (20 in Llama, 6 in Gemma, 25 in Ministral). Scaling is evaluated on 400 \textit{test-clean} and 400 \textit{test-other} utterances not used for blocking. We use five audio conditions: clean audio, and babble and white noise at 5 and 0\,dB SNR.
\subsection{Results}
\setlength{\intextsep}{4pt plus 1pt minus 1pt}
\renewcommand{\arraystretch}{0.9}
WER under attention head blocking is reported for each decoder in Table~\ref{tab:rq1}. Blocking $\mathcal{C}_{10}$ increases WER in all three decoders. The increase is much larger in Gemma, where 10 heads are about 5\,\% of all heads, against 1.5 and 1.2\,\% in Llama and Ministral, and varies heavily between seeds. In Llama and Ministral it is milder, yet still above that of random sets of heads of the same size. Moreover, $\mathcal{C}_k$ makes up 40, 33 and 40\,\% of the 3\,\% of heads with the largest ASR importance in Llama, Gemma and Ministral, against 4--9\,\% for random heads of the same layers. Within the speech embeddings, the attention of $\mathcal{C}_k$ peaks on the word being generated for 64--76,\% of tokens, against 45--56,\% for equally many heads with the largest attention mass on $S$. These results show that ASR in SLAM-ASR models is concentrated in induction heads of the frozen decoder.
\begin{table}[t]
\centering
\vspace{-8pt}
\caption{WER (\%) when the attention logits of $\mathcal{C}_{10}$ on $S$ are set to $-\infty$; random: mean WER over 10 sets of 10 heads from the same layers, blocked in the same way. Ranges are over 3 seeds throughout.}
\label{tab:rq1}
\small
\setlength{\tabcolsep}{4pt}
\begin{tabular*}{\columnwidth}{@{\extracolsep{\fill}}lccc@{}}
\toprule
decoder & intact & $\mathcal{C}_{10}$ blocked & random \\
\midrule
Llama     & 3.3--3.6 & \textbf{7.1--8.3}   & 3.3--3.6 \\
Gemma     & 3.8--4.0 & \textbf{23.2--53.5} & 4.0--4.2 \\
Ministral & 3.0--3.4 & \textbf{3.9--5.5}   & 3.1--3.4 \\
\bottomrule
\end{tabular*}
\end{table}
\begin{table}[b]
\centering
\vspace{-8pt}
\caption{Attention noise cancelling head pairs, given as layer: $h_1$, $h_2$. The pair is the same in all three seeds. $-\cos$: negated median cosine between $W_O^{h_1}v_j$ and $W_O^{h_2}v_j$, $j\in S$; $\lambda$: scaling factor between the two; $\rho$: share of positive attention on speech embeddings of the previously generated word, for $h_1$ alone and for the pair (eff.); WER (\%) with attention to speech blocked for $h_1$ alone or both.}
\label{tab:rq2}
\footnotesize
\setlength{\tabcolsep}{2pt}
\newcommand{\pair}[2]{\begin{tabular}[c]{@{}l@{}}#1\\[-1pt]\scriptsize #2\end{tabular}}
\begin{tabular*}{\columnwidth}{@{\extracolsep{\fill}}lcccccc@{}}
\toprule
 & & & \multicolumn{2}{c}{$\rho$} & \multicolumn{2}{c}{WER, blocked} \\
\cmidrule(lr){4-5}\cmidrule(lr){6-7}
\pair{decoder}{layer: $h_1,h_2$} & $-\cos$ & $\lambda$ & $h_1$ & eff. & $h_1$ & both \\
\midrule
\pair{Llama}{8: H8, H6}     & 0.72--0.74 & 0.78--0.82 & 0.52--0.57 & \textbf{0.64--0.67} & 45--56    & 3.8--7.2 \\
\pair{Gemma}{8: H1, H0}     & 0.66--0.67 & 0.92--0.94 & 0.74--0.76 & \textbf{0.80--0.81} & 58--149   & 15--39 \\
\pair{Ministral}{14: H0, H2} & 0.78--0.79 & 0.73--0.74 & 0.78--0.80 & \textbf{0.80--0.81} & 5.1--13.8 & 3.3--4.2 \\
\bottomrule
\end{tabular*}
\end{table}

Head pairs that meet the conditions of an attention noise cancelling pair are reported for each decoder in Table~\ref{tab:rq2}. In all three decoders, the $h_2$ output for each speech embedding is approximately a negative multiple of the $h_1$ output, as the $-\cos$ column shows, indicating that the two heads act as one head with \mbox{differential} attention~\cite{ye2025}. The effective attention of the pairs also supports this view: as the eff.\ column shows, it is sharper on the speech embeddings of the previously generated word than the attention of $h_1$ alone in all decoders, though only slightly in Ministral. Blocking $h_1$ alone increases WER far more than blocking both heads, whereas blocking $h_2$ alone leaves it unchanged in Gemma and Ministral: the two heads act as a unit, and blocking one leaves the output of the other uncancelled. Blocked together, the pair costs little in Llama and Ministral and more in Gemma; its sharper attention describes how the pair reads the speech rather than how much recognition depends on it.

WER and insertion rate $I$ under scaling are reported for each decoder in Table~\ref{tab:rq3}. Under noisy audio, scaling $\mathcal{C}_k$ results in lower WER and $I$ than scaling the same number of random heads (by 1.4--2.6 WER in the noisy mean) or the remaining heads of the same layers (by 4.6--10.9) in all three decoders, which indicates that the reduction is not an effect of scaling heads in these layers as such. Scaling instead equally many heads with the largest ASR importance reduces WER less than scaling $\mathcal{C}_k$ in Llama (by 2.4, noisy mean), about as much in Ministral, and more in Gemma (by 4.5). Compared to the unscaled model, scaling $\mathcal{C}_k$ reduces WER and $I$ in Llama and Ministral almost only under babble noise at 0\,dB, where insertions are most frequent; under the other noisy conditions WER changes by less than 1 point. In Gemma, scaling heads of these layers increases WER, and scaling $\mathcal{C}_k$ increases it least. On clean audio, scaling $\mathcal{C}_k$ increases WER by at most 0.2 points.
\begin{table}[t]
\centering
\vspace{-8pt}
\caption{WER and $I$ (\%) with heads scaled by $\alpha=1.5$: none (unscaled model), $\mathcal{C}_k$ ($k$ = 3\,\% of heads), equally many random heads from the same layers (mean over 10 sets), or all remaining heads of the same layers. Clean: WER without added noise on 400 \textit{test-clean} and 400 \textit{test-other} utterances; noisy (mean): over babble and white noise at 5 and 0\,dB SNR added to the same utterances. Means over 3 seeds; in each seed, $\mathcal{C}_k$ gives lower noisy WER and $I$ than random heads. Bold: lowest per decoder under noise.}
\label{tab:rq3}
\footnotesize
\setlength{\tabcolsep}{3pt}
\begin{tabular*}{\columnwidth}{@{\extracolsep{\fill}}llccccc@{}}
\toprule
 & & & \multicolumn{2}{c}{babble 0\,dB} & \multicolumn{2}{c}{noisy (mean)} \\
\cmidrule(lr){4-5}\cmidrule(lr){6-7}
decoder & scaled & clean & WER & $I$ & WER & $I$ \\
\midrule
Llama     & none              & 4.6 & 89.0           & 40.4          & 41.9          & 15.0 \\
          & $\mathcal{C}_k$   & 4.7 & \textbf{81.8}  & \textbf{32.1} & \textbf{39.9} & \textbf{12.4} \\
          & random            & 4.5 & 89.9           & 41.3          & 42.4          & 15.5 \\
          & same layers       & 4.8 & 111.8          & 63.6          & 50.8          & 23.8 \\
\midrule
Gemma     & none              & 4.7 & \textbf{148.6} & 94.9          & \textbf{66.9} & 37.9 \\
          & $\mathcal{C}_k$   & 4.8 & 150.0          & \textbf{94.7} & 67.4          & \textbf{37.7} \\
          & random            & 4.8 & 153.6          & 99.0          & 70.0          & 40.7 \\
          & same layers       & 5.2 & 156.2          & 101.5         & 72.0          & 42.6 \\
\midrule
Ministral & none              & 4.2 & 86.0           & 39.3          & 39.6          & 14.1 \\
          & $\mathcal{C}_k$   & 4.4 & \textbf{81.8}  & \textbf{34.4} & \textbf{38.5} & \textbf{12.6} \\
          & random            & 4.3 & 86.0           & 39.6          & 39.9          & 14.4 \\
          & same layers       & 4.6 & 94.0           & 45.6          & 43.4          & 16.6 \\
\bottomrule
\end{tabular*}
\end{table}
\subsection{Discussion}
The link between SLAM-ASR models and induction heads of the frozen LLM gives a concrete answer to the question we posed in Sec.~\ref{sec:intro}: it suggests that a small map suffices because its outputs only need to be readable by the LLM's existing context-reading circuits, which perform prefix matching on them. This raises the empirical question of how projector training should be done, especially when done jointly with the decoder. For instance, updating only the heads with the highest $\mathrm{PM}$ score, rather than the whole decoder, is a potentially promising approach.

The attention noise cancelling pairs we found show that text-pretrained LLMs contain heads that act together to cancel attention noise on redundant input, without being explicitly trained for it, while~\cite{dex} has to add differential attention to a pretrained LLM through adaptation. Heads that read the same input and project in opposite directions are known in text LLMs as negative and copy-suppression heads~\cite{wang2023ioi,mcdougall2024}, which act on the logits of copied tokens; the pairs we found share keys and values, and their subtraction acts on attention. It is not clear how these heads emerge during text pretraining, which is a promising direction for mechanistic interpretability research.

\section{Conclusion}
\label{sec:conclusions}
In this work, we have shown that speech recognition with LLMs in models such as SLAM-ASR depends on induction heads of the frozen LLM. In the decoders we considered, pairs of attention heads act together as a constrained form of differential attention, cancelling attention noise and sharpening attention on the speech embeddings of the previously generated word. Induction heads also matter when doing recognition on noisy audio: scaling them at inference mitigates hallucinations in two of the three decoders we consider.

\bibliographystyle{IEEEbib}
\bibliography{refs}

\end{document}